\documentclass[10pt,twocolumn,letterpaper]{article}
\usepackage[pagenumbers]{iccv} 

\usepackage{graphicx}
\usepackage{amsmath}
\usepackage{amssymb}
\usepackage{booktabs}
\usepackage{soul}
\usepackage{multirow}
\usepackage{diagbox}
\usepackage{tabularx}
\usepackage{xr}
\definecolor{iccvblue}{rgb}{0.21,0.49,0.74}
\usepackage[pagebackref,breaklinks,colorlinks,allcolors=iccvblue]{hyperref}

\title{Tiling artifacts and trade-offs of feature normalization in the segmentation of large biological images}

\author{Elena Buglakova$^1$, Anwai Archit$^2$, Edoardo D'Imprima$^{1,4}$, Julia Mahamid$^1$, \\ Constantin Pape$^{2, 3}$, Anna Kreshuk$^1$\\
$^1$  {\small European Molecular Biology Laboratory, Heidelberg} \\
$^2$  {\small Institute of Computer Science, University of Göttingen} \\
$^3$  {\small Cluster of Excellence ‘Multiscale Bioimaging: from Molecular Machines to Networks of Excitable Cells‘} \\ {\small (MBExC), Georg-August-University Göttingen} \\
$^4$  {\small IRCCS Humanitas Research Hospital, Milan}
}

\begin{document}
\maketitle
\begin{abstract}
Segmentation of very large images is a common problem in microscopy, medical imaging or remote sensing. The problem is usually addressed by sliding window inference, which can theoretically lead to seamlessly stitched predictions. However, in practice many of the popular pipelines still suffer from tiling artifacts. We investigate the root cause of these issues and show that they stem from the normalization layers within the neural networks. We propose indicators to detect normalization issues and further explore the trade-offs between artifact-free and high-quality predictions, using three diverse microscopy datasets as examples. Finally, we propose to use BatchRenorm as the most suitable normalization strategy, which effectively removes tiling artifacts and enhances transfer performance, thereby improving the reusability of trained networks for new datasets. 
\end{abstract}    
\section{Introduction}
\label{sec:intro}

Segmentation is an essential first step of pipelines in biological image analysis, a diverse domain of computer vision that enables automated quantification for various types of microscopy images. While this domain has benefited a lot from the developments in natural image analysis, it also poses unique challenges. Investigation of biological diversity, discovery of unexpected phenotypes and development of novel imaging protocols leads to high variability of datasets. This explains the popularity of human-in-the-loop approaches to segmentation, where a small amount of data is either annotated from scratch or from the output of pre-trained models \cite{Stringer2020, Archit2023, Weigert2020}, then a new model is either trained from scratch or fine-tuned. After that the next rounds of correction and training happen, until the desired quality is reached. Being able to quickly train a smaller model for an unseen image type is crucial in this setting.

Another important feature of biological images, especially for volumetric imaging, is their large size. Electron microscopy volumes for connectomics \cite{Dorkenwald2024} or light sheet microscopy volumes \cite{Ruan2024} can reach the scale of multiple terabytes. It is routine to work with 10-100 GB images of $\sim 1000^3$ pixels, which exceed  GPU memory. 
\begin{figure}[t]
  \centering
    \includegraphics[width=\linewidth]{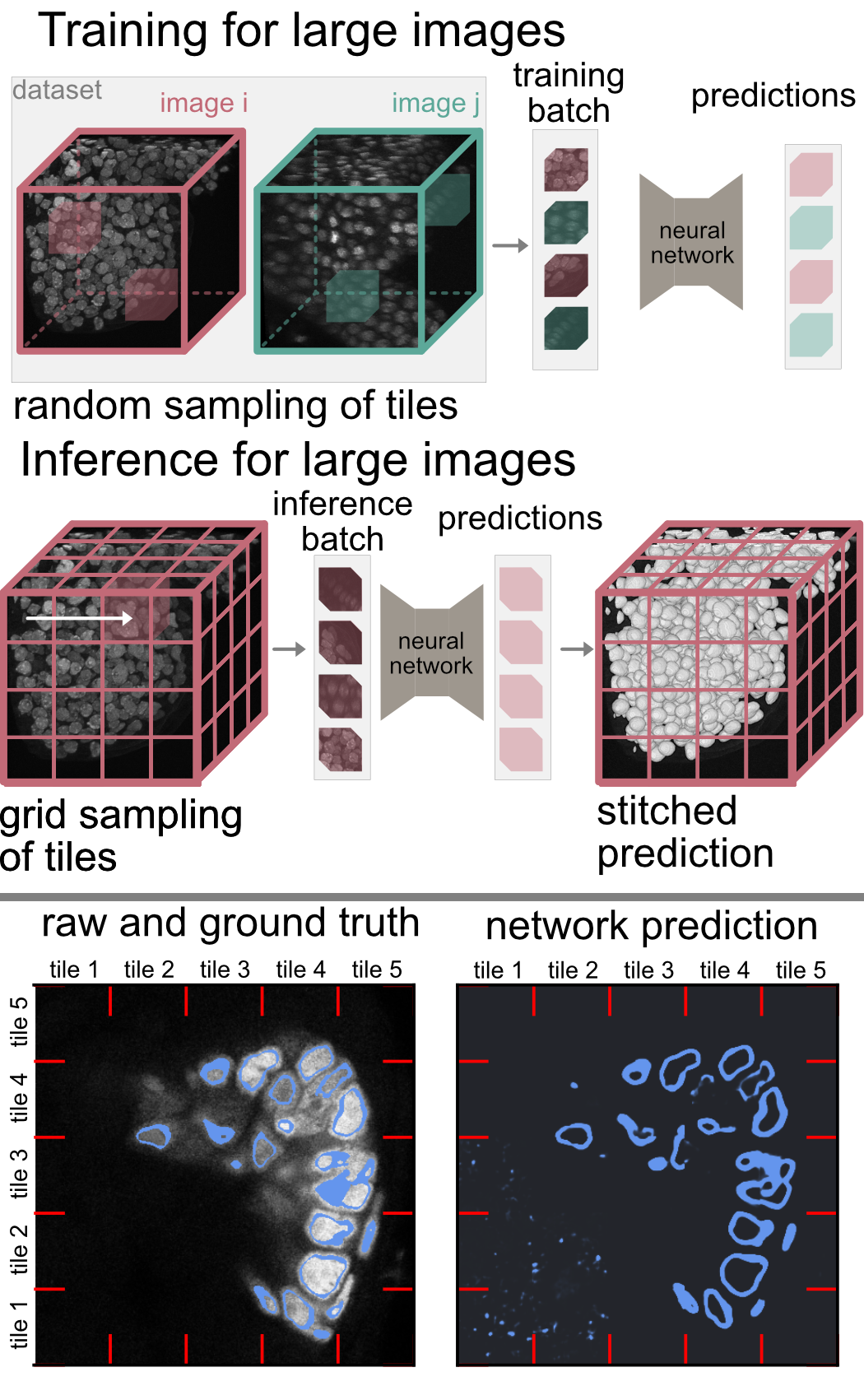}

   \caption{ Illustration of the pipeline for processing images larger than GPU memory: random sampling of tiles during training and sliding window inference. Example of artifacts caused by tiling: hallucinations in the low signal areas and discontinuous predictions at the tile borders.
   }
   \label{fig:pipeline}
\end{figure}

\begin{figure*}[t]
  \centering
    \includegraphics[width=\linewidth]{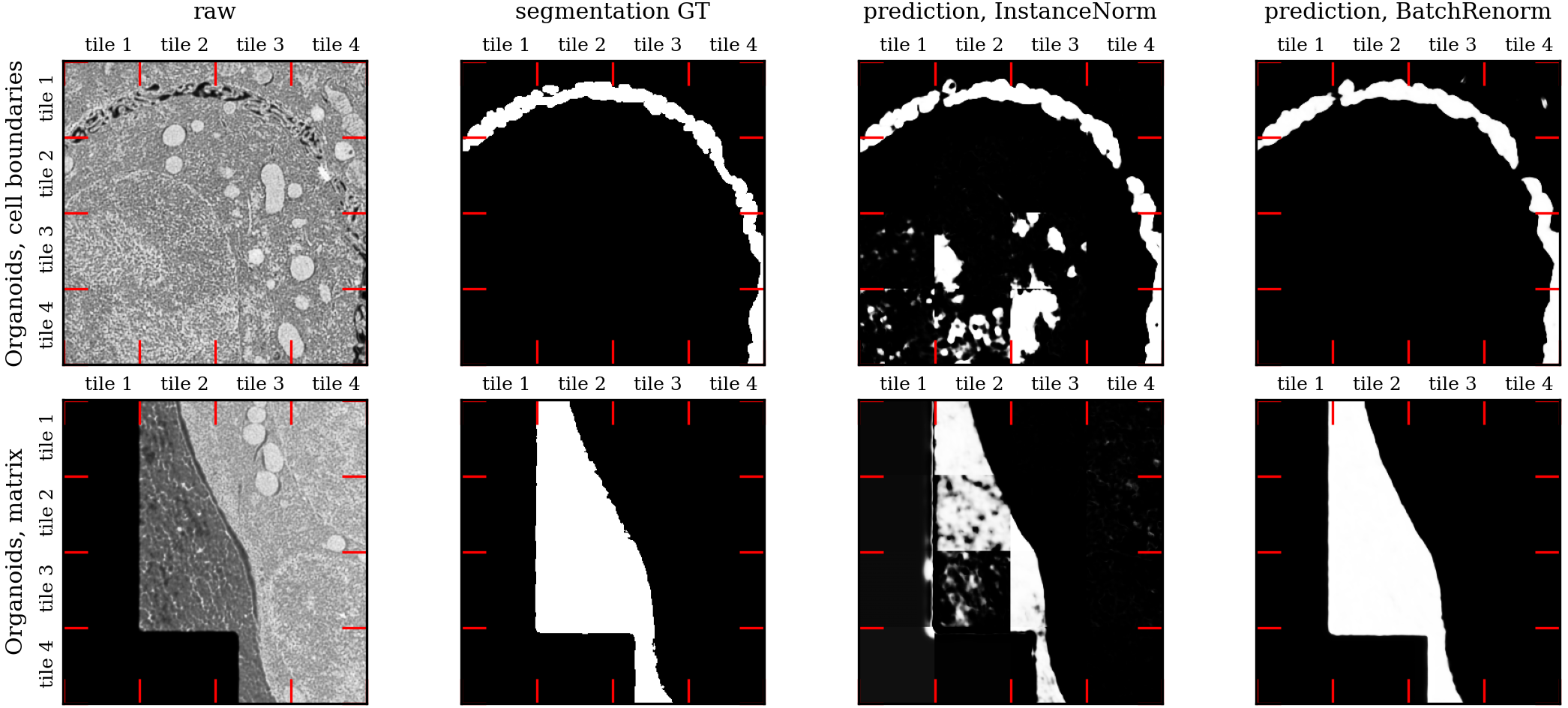}

   \caption{Examples of predictions with and without tiling artifacts}
   \label{fig:artifact_correction}
\end{figure*}

As shown in \cref{fig:pipeline}, during training random patches are sampled from the annotated images and combined into a batch. During inference/evaluation, the image is split into a grid of tiles which are predicted independently and then stitched back to form the full prediction. Here we refer both to 2D and 3D image patches as tiles. In this setup the size of the tiles is limited from above by GPU memory and from below by the receptive field of the network \cite{Rumberger2021}. Typical tile size is around 96x96x96 - 256x256x256 pixels and the training batch size is 2-3 tiles per batch. During inference it is not necessary to store gradients, therefore either tile size or batch size can be larger. 

Tile-wise prediction and stitching might look like a purely technical step, however for microscopy patch sampling and stitching can significantly affect prediction quality and create artifacts. In contrast, big natural image datasets, such as ImageNet \cite{ImageNet}, CityScapes \cite{Cordts2016Cityscapes} and SAM dataset \cite{SAM}, usually contain relatively small 2D images, making tiling and stitching unnecessary, as well as allowing for larger batch size.

It has been shown that seamless stitching with CNNs is possible if edge effects are taken into account \cite{Majurski2021}. For CNNs only a limited area of the input image called the receptive field affects the prediction value in a given pixel \cite{NIPS2016_c8067ad1}. Pixels on the edge of the tile do not get full context for prediction, resulting in a mismatch between neighboring tiles. To deal with this issue the tiles are sampled with overlap and only the central part of each tile is used for stitching. We refer to the removed area as halo. We find that even with sufficient halo sliding window inference can cause tiling artifacts manifesting in two ways: as abrupt discrepancy between predictions in neighboring tiles or as background hallucinations, as shown in \cref{fig:pipeline} and \cref{fig:artifact_correction}. This problem is rarely discussed but it affects many widely used tools, including nnU-Net \cite{Isensee2020}, a popular baseline method for biomedical image segmentation. 

A common way to deal with the tiling artifacts is to apply additional postprocessing, for example, to average neighboring tiles with gaussian weights to avoid discontinuous predictions \cite{Isensee2020}, or to use additional thresholding by intensity and filtering by the object size to remove artifacts in the background like in the \cref{fig:pipeline}. However, adding heuristic postprocessing steps makes the whole pipeline harder to use and apply to new datasets. Moreover, we show that tiling artifacts become more pronounced in the transfer setting, further reducing reusability of the trained networks.

According to our findings, the main cause of the artifacts is the tile-wise feature normalization inside the network. If \texttt{InstanceNorm} \cite{InstanceNorm} is used in the CNN architecture, both during training and evaluation the statistics of the current input are used for normalization. This makes predictions for every pixel dependent on the content of the whole tile, therefore the limited receptive field assumption becomes invalid and, regardless of the size of the halo, it is impossible to make the stitching seamless.

Unlike \texttt{InstanceNorm}, another popular normalization method \texttt{BatchNorm} \cite{pmlr-v37-ioffe15} uses the statistics of the current input during training but switches to saved running average statistics in the evaluation mode, allowing to achieve seamless stitching. However, we show that the performance of the network drastically reduces in evaluation mode compared to training mode because the statistics used during training are not stable with small batch size, making the running average not representative of the statistics of individual batches. This effect has previously been observed for natural image classification \cite{NormReview, BatchRenorm}, but we find it to be significantly more pronounced in biological image segmentation.

In this work we demonstrate the trade-off between the artifact-free stitching ensured by the global normalization with \texttt{BatchNorm} and prediction quality achieved by tile-wise normalization with \texttt{InstanceNorm}. We suggest \texttt{BatchRenorm} \cite{BatchRenorm} which uses the same global normalization statistics both during training and inference as a solution which produces artifact-free predictions without compromising on the prediction quality.

In summary, we make the following contributions:
\begin{itemize}
\item We show the existence of a normalization trade-off in segmentation of large biological images: using tile-wise statistics for feature normalization in the network during inference (\texttt{InstanceNorm}) leads to tiling artifacts even if edge effects were taken into consideration. Using tile-wise statistics in training and global accumulated statistics during inference (\texttt{BatchNorm}) leads to an unexpected difference between network performance during training and inference due to small batch size and high variability of the data;
\item We suggest metrics for detecting normalization issues in networks used for sliding window inference pipelines: \textbf{tile mismatch} and \textbf{train/eval disparity};
\item We propose to use the \textbf{Batch Renormalization} method to achieve seamless stitching without sacrificing prediction accuracy to the train/eval disparity;
\item We validate our findings on three volumetric biological datasets, with two network architectures: CNN-based 3D U-Net and ViT-based UNETR.
\end{itemize}

\section{Related work}
\label{sec:related_work}
\begin{figure}[t]
  \centering

    \includegraphics[width=\linewidth]{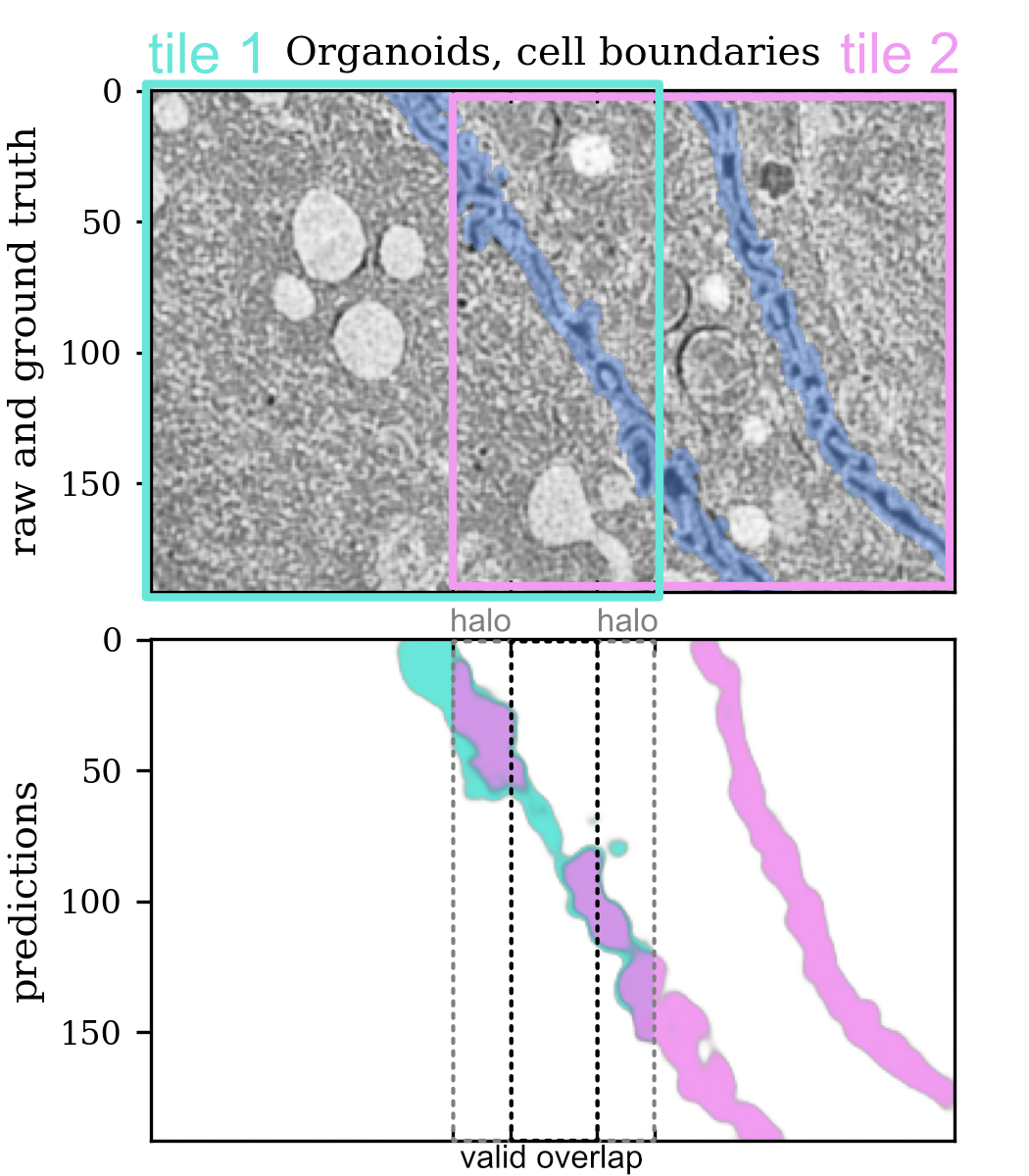}

   \caption{Illustration of the tile mismatch metric.}
   \label{fig:tile_mismatch}
\end{figure}

\begin{figure*}[t]
  \centering

    \includegraphics[width=\linewidth]{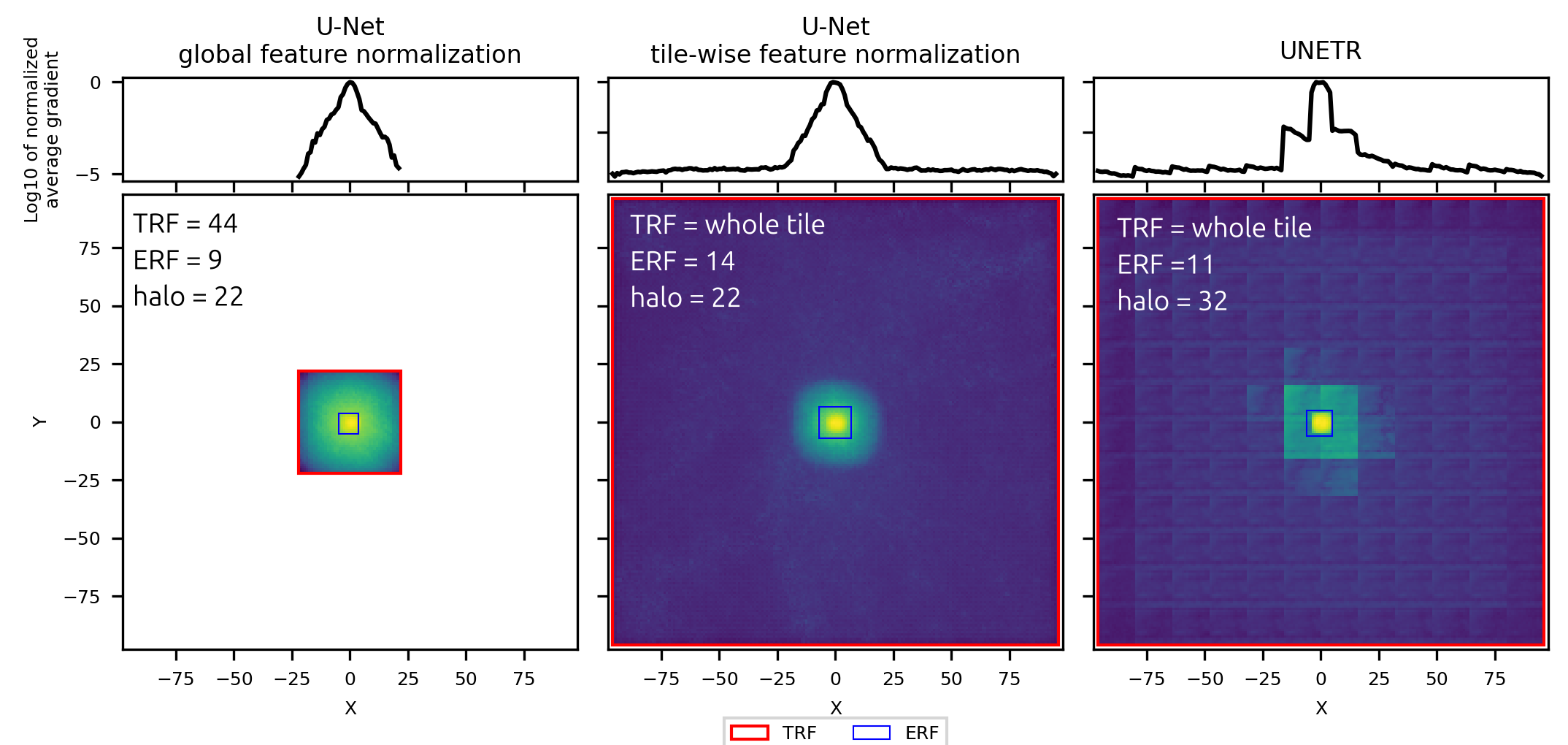}

   \caption{Receptive field, shown as Log10 of mean gradient of the central pixel output with respect to the input}
   \label{fig:receptive_field}
\end{figure*}

Sliding window inference is used in different domains, including medical \cite{Ronneberger2015, iek2016}, biological \cite{FMBCAMBBR19} and aerial \cite{Huang2018TilingAS} image analysis. It was shown that receptive field ($RF$) \cite{NIPS2016_c8067ad1, Loos2024} plays a key role in the seamless stitching \cite{Majurski2021}. Often in case of dense prediction the input and output have the same size thanks to the zero-padding in convolutional layers, so the coordinates in both spaces can be used interchangeably. For the output pixels which are closer to the edge of the tile than $\frac{1}{2}RF$ part of the input values within the receptive field is replaced by zeros, creating mismatch in the predictions between tiles. Proposed solution is to crop the prediction and only use the part of the output where each pixel has full context for stitching. The minimal sufficient size of the cropped area is $\frac{1}{2}RF$. The same concept was discussed from the point of view of translational equivariance \cite{Huang2018TilingAS, Rumberger2021}. Although convolutions and pooling are translation-equivariant, zero-padding makes the CNN overall lose this property. With valid convolutions the size of the output is smaller than the input but it allows to keep the translational equivariance and therefore seamless stitching.     

In practice many pipelines using sliding window inference do not aim to completely eliminate the tiling artifacts but rather minimize them by using the largest possible tile size in inference and averaging. For example, nnU-Net \cite{Isensee2020} and Cellpose \cite{Stringer2020} compute weighted average of the tiles, with the weight decreasing from 1 in the middle of the tile to 0 at the border, smoothing the border between tiles. 

These recommendations help against the discontinuity in predictions caused by edge effects but our experiments show that even in the valid convolution area predictions do not necessarily match (see Fig.~\ref{fig:tile_mismatch}). In addition, this does not explain hallucination-like artifacts in the background like in \cref{fig:pipeline}. It has been shown that prediction accuracy can depend significantly on the tile size \cite{Reina2020, Cira2024}. To our knowledge, the connection between these effects and feature normalization inside the network was not studied before.

\section{Methods}
\label{sec:methods}
This section is structured as follows: we define how  we perform sliding window inference and which normalization strategies we consider. We then propose two metrics that can identify normalization issues within the network, pointing out both tiling artifacts ("tile mismatch" metric) and performance drop in evaluation mode ("train/eval disparity" metric).

\subsection{Sliding window inference with halo}
Input data can be a 2D image or a 3D volume with one or multiple channels. Here both images and volumes are referred to as images. For inference the image is split into a grid of tiles, incomplete tiles are padded with zeros. The stitched prediction is calculated from by tile-wise predictions.  To ensure that observed tiling artifacts were not caused by edge effects, we calculated theoretical (TRF) and effective receptive field (ERF) of the networks following \cite{Loos2024}. We define TRF as all pixels which belong to the computation graph for the target pixel in the output space. During sliding window inference the overlap was set to the size of halo, minimizing the number of tiles needed to cover the whole volume. The halo part of the tile was completely removed so there was no averaging of predictions or other postprocessing.

\subsection{Normalization strategies}

Normalization can be inserted in two parts of the pipeline: preprocessing of the input data and feature map normalization inside the network.

\subsubsection{Input normalization} 

Quantile normalization with $q_{min}=0.01$ and $q_{max}=0.98$ and clipping to $[0..1]$ were applied to each image as a whole to convert the data type and value range from the microscope output to $float32$ and $[0..1]$ range. In addition to the global normalization, quantile normalization can be applied to each tile separately to make the input of the network always have the same range and make tiles more comparable. Based on this we considered two strategies:

\begin{itemize}
\item \textit{Global}: quantile normalization only of the whole image.

\item \textit{Tile-wise}: quantile normalization of each sampled tile, both during training and during inference.

\end{itemize}
\subsubsection{Feature normalization}  
 General formula for the normalization operation with input $x$, output $y$ is:

$$
y = \frac{x-\mu}{\sqrt{\sigma^2 + \epsilon}},
$$
where $\mu$ and $\sigma$ are normalization parameters and $\epsilon$ is a small constant used for numerical stability. Parameters $\mu$ and $\sigma$ can be estimated directly from the input $\mu = \mathrm{E}[x]$ and $\sigma^2 = \mathrm{Var}[x]$, where average can be taken either over each sample independently or over the whole batch. Alternatively, global normalization parameters independent of the current input can be used. A common strategy is to estimate the parameters as a running average over multiple samples: $p_{new} = (1 - momentum) \times p_{old} + momentum \times p_{t}$. The update speed is determined by the $momentum$.

\begin{itemize}

\item \texttt{BatchNorm} \cite{pmlr-v37-ioffe15} Training: use statistics of the current batch and collect running average. Inference: use saved running average.

\item \texttt{InstanceNorm} \cite{InstanceNorm} Both training and inference: use statistics of each input sample. In Pytorch this behavior can be changed by setting \texttt{track\_running\_stats=True}. Then during inference the statistics collected during training will be utilized, same as in \texttt{BatchNorm}.

\item \texttt{BatchRenorm} \cite{BatchRenorm} Both training and inference: use running average statistics.

\item \texttt{Identity} Skip feature normalization.

\subsection{Proposed evaluation metrics}

\textbf{Train/eval disparity}
When global normalization with running average statistics is used during inference, a disparity can arise between the predictions of the same tile in train and eval mode as the train mode always uses statistics of the current batch. This disparity is an indicator of a potential drop in performance which is not related to model overfitting. To measure it quantitatively, we introduce the following metric: 
$$
train/eval\ disparity = 1 - Dice(P_{train}, P_{eval}), 
$$ 
where $P_{train}$ is the prediction done with  \texttt{model.train()} and $P_{eval}$ - with \texttt{model.eval()}. 

\textbf{Tile mismatch}

Quantification of tiling artifacts is challenging because their magnitude depends on the similarity of the content of neighboring tiles in different parts of the image. We propose to split the image into overlapping tiles and compare the predictions in the overlap areas. To avoid edge effects, we only take the valid part of the overlap, as shown in \cref{fig:tile_mismatch}.  The following metrics are proposed:

$$
max\ dist = \max_{i=1}^M(|O_{i1} - O_{i2}|) 
$$

$$
tile\ mismatch = \mathop{\mathit{median}}_{i=1}^{M}(1 - Dice(O_{i1}, O_{i2})), 
$$

where $O_{i1}$ and $O_{i2}$ are predictions in the valid overlap region in tile $i$ and $M$ is number of sampled tiles. In $max\ dist$ $\max$ is taken over all channels and all tiles, making it an indicator of wether given setup produces tiling artifacts. In cases when $max\ dist = 0$ we explicitly report "no" instead of $tile\ mismatch$ to emphasize that the predictions perfectly match.  $tile\ mismatch$ is calculated per channel and characterizes the magnitude of the artifacts. 

\end{itemize}
\begin{table*}
\small
\centering
\caption{U-Net results table}
\label{tab:unet}

\begin{tabular}{c|l|l||l|l|l|l|l|l}

\toprule
\multicolumn{3}{l||}{Normalization layer} & BatchNorm & BatchNorm & InstanceNorm & InstanceNorm & BatchRenorm & Identity \\

\multicolumn{3}{l||}{Input norm} & global & tile-wise & global & tile-wise & global & global  \\

\multicolumn{3}{l||}{Feature norm} & global & global & tile-wise & tile-wise & global & global  \\
\midrule
\midrule
\multirow{7}{*}{\rotatebox[origin=c]{90}{dice, eval mode}} & \multirow{3}{*}{organoids}
& foreground & \textbf{0.95} & 0.88 & 0.94 & 0.93 & \textbf{0.95} & \textbf{0.95} \\
& & boundaries & 0.49 & 0.15 & \textbf{0.75} & \textbf{0.75} & 0.73 & 0.72 \\
& & matrix & 0.82 & 0.61 & 0.61 & 0.61 & \textbf{0.85} & 0.83 \\
\cline{2-9}
& \multirow{2}{*}{plants}
 & foreground & 0.85 & 0.88 & \textbf{0.89} & \textbf{0.89} & \textbf{0.89} & 0.52 \\
& & boundaries & 0.69 & 0.73 & \textbf{0.75} & 0.74 & \textbf{0.75} & 0.26 \\
\cline{2-9}
& \multirow{2}{*}{embryo}
 & foreground & 0.66 & 0.68 & 0.67 & 0.72 & \textbf{0.74} & 0.59 \\
& & boundaries & 0.55 & 0.55 & 0.55 & 0.53 & \textbf{0.58} & 0.44 \\
\midrule
\midrule
\multirow{7}{*}{\rotatebox[origin=c]{90}{dice, train mode}} & \multirow{3}{*}{organoids}
& foreground & 0.94 & 0.93 & 0.94 & 0.93 & \textbf{0.95} & 0.95 \\
& & boundaries & \textbf{0.75} & \textbf{0.75} & \textbf{0.75} & \textbf{0.75} & 0.73 & 0.72 \\
& & matrix & 0.61 & 0.61 & 0.61 & 0.61 & \textbf{0.85} & 0.83 \\
\cline{2-9}
& \multirow{2}{*}{plants}
 & foreground & \textbf{0.89} & \textbf{0.89} & \textbf{0.89} & \textbf{0.89} & \textbf{0.89} & 0.52 \\
& & boundaries & \textbf{0.75} & 0.74 & \textbf{0.75} & 0.74 & \textbf{0.75} & 0.26 \\
\cline{2-9}
& \multirow{2}{*}{embryo}
 & foreground & 0.67 & 0.72 & 0.67 & 0.72 & \textbf{0.74} & 0.59 \\
& & boundaries & 0.55 & 0.52 & 0.55 & 0.53 & \textbf{0.58} & 0.44 \\

\midrule
\midrule
\multicolumn{2}{l|}{\multirow{3}{*}{train/eval disparity}}
& organoids & 0.48 & 0.84 & \textbf{0} & \textbf{0} & \textbf{0} & \textbf{0} \\
\cline{3-9}
\multicolumn{2}{l|}{\multirow{3}{*}{}} & plants & 0.17 & 0.11 & \textbf{0} & \textbf{0} & \textbf{0} & \textbf{0} \\
\cline{3-9}
\multicolumn{2}{l|}{\multirow{3}{*}{}} & embryo &  0.37 & 0.29 & \textbf{0} & \textbf{0} & \textbf{0} & \textbf{0} \\

\midrule
\midrule
\multicolumn{2}{l|}{\multirow{3}{*}{tile mismatch}} 
& organoids & \textbf{no} & 0.03 & 0.11 & 0.11 & \textbf{no} & \textbf{no} \\
\cline{3-9}
\multicolumn{2}{l|}{\multirow{3}{*}{}} &  plants & \textbf{no} & 0.01 & 0.05 & 0.05 & \textbf{no} & \textbf{no} \\
\cline{3-9}
\multicolumn{2}{l|}{\multirow{3}{*}{}} & embryo & \textbf{no} & 0.09 & 0.16 & 0.20 & \textbf{no} & \textbf{no} \\

\midrule
\bottomrule
\end{tabular}
\end{table*}

\section{Experiments}
\label{sec:experiments}

\subsection{Datasets}

The datasets were chosen to demonstrate the common challenges of microscopy  segmentation such as large image size and variability of class distributions within the volume. 

\textbf{organoids} \cite{Dimprima2023}: semantic segmentation of tissue in patient-derived colorectal cancer organoids, electron microscopy (FIB/SEM), $40 \times 30 \times 30$ nm$^3$ voxel size. The volume of size $1350 \times 1506 \times 1647$ pixels was split 70/30 into train and validation dataset. Classes: cell boundaries, foreground (pixels inside the cells), extracellular matrix (space between cells  and background).

\textbf{plants} \cite{Vijayan2024}: instance segmentation of nuclei in Arabidopsis ovules, confocal fluorescent microscopy, isotropic resolution of $0.13\ \mu m^3$ per pixel, 4 train volumes and 1 validation volume $\sim 500 \times 1000 \times 1000$ pixels each. 

\textbf{embryo} \cite{Bondarenko2023}: instance segmentation of nuclei in mouse embryos, confocal fluorescent microscopy, isotropic resolution of $0.2\ \mu m^3$ per pixel, 22 train volumes and 13 validation volumes, $\sim 600 \times 1000 \times 700$ px each. 

For \textbf{plants} and \textbf{embryo} datasets the original instance segmentation groundtruth was converted to semantic segmentation of three classes: object boundaries, foreground (objects with subtracted boundaries) and background. For the \textbf{transfer setup}, networks trained for \textbf{embryo} dataset were used to segment \textbf{plants} dataset and vice versa.

\subsection{Architectures}

\textbf{3D U-Net}: 3D U-Net with 2 downsampling steps and 2 convolutional blocks per level was used for all datasets. Each convolutional block consists of a convolutional layer, a normalization layer and a non-linearity. The normalization layers in all the blocks were replaced with one of the normalization layers listed above. Tile size $(192, 192, 192)$ was used for training and $(128, 128, 128)$ for sliding window inference. 

\textbf{3D UNETR}: UNETR consists of a ViT encoder and a convolutional decoder. In the encoder the volume is split into $16 \times 16 \times 16$ patches and embeddings in the ViT are normalized using \texttt{LayerNorm} after each round of attention. With \texttt{LayerNorm} all features are normalized using average and variance of the features in each patch independently. Other normalization strategies can be used in the ViT \cite{Yao2021}, but here \texttt{LayerNorm} was kept in all experiments as a standard approach. Convolutional decoder consists of 4 upsampling steps with 2 convolutional blocks at each level of the U-Net-like architecture. The normalization layers in all the blocks were replaced with one of the normalization layers listed above. 

\subsection{Reported metrics}
\textbf{Overall prediction quality}: Prediction was done on the whole volume using sliding window inference as described above. If the setup included tile-wise input normalization, each tile was normalized separately before being passed through the network. The median Dice score over all samples in the validation dataset was reported for each class. 
\textbf{Train/eval disparity}: We report the Dice score both for training and evaluation mode to demonstrate the disparity in \cref{tab:unet} and \cref{tab:unetr}. Inference in both modes was done on the same images from the validation set, so the difference comes only from the behavior of the model in different modes and not overfitting to the training data.
\textbf{Tile mismatch}: The tiles of $192 \times 192 \times 288$px were sampled in a grid with a stride of 64 pixels to test the mismatch in the whole volume. Sampled tiles were split into two overlapping tiles of size $192 \times 192 \times 192$. Both tiles were then processed to compare the predictions in the overlapping region. 
Tile mismatch values for boundary channel are reported in \cref{tab:unet}, \cref{tab:unetr} and STab.~1.

Further details can be found in the  Supplementary.

\section{Results}
\label{sec:results}
\begin{figure}[t]
  \centering
    \includegraphics[width=\linewidth]{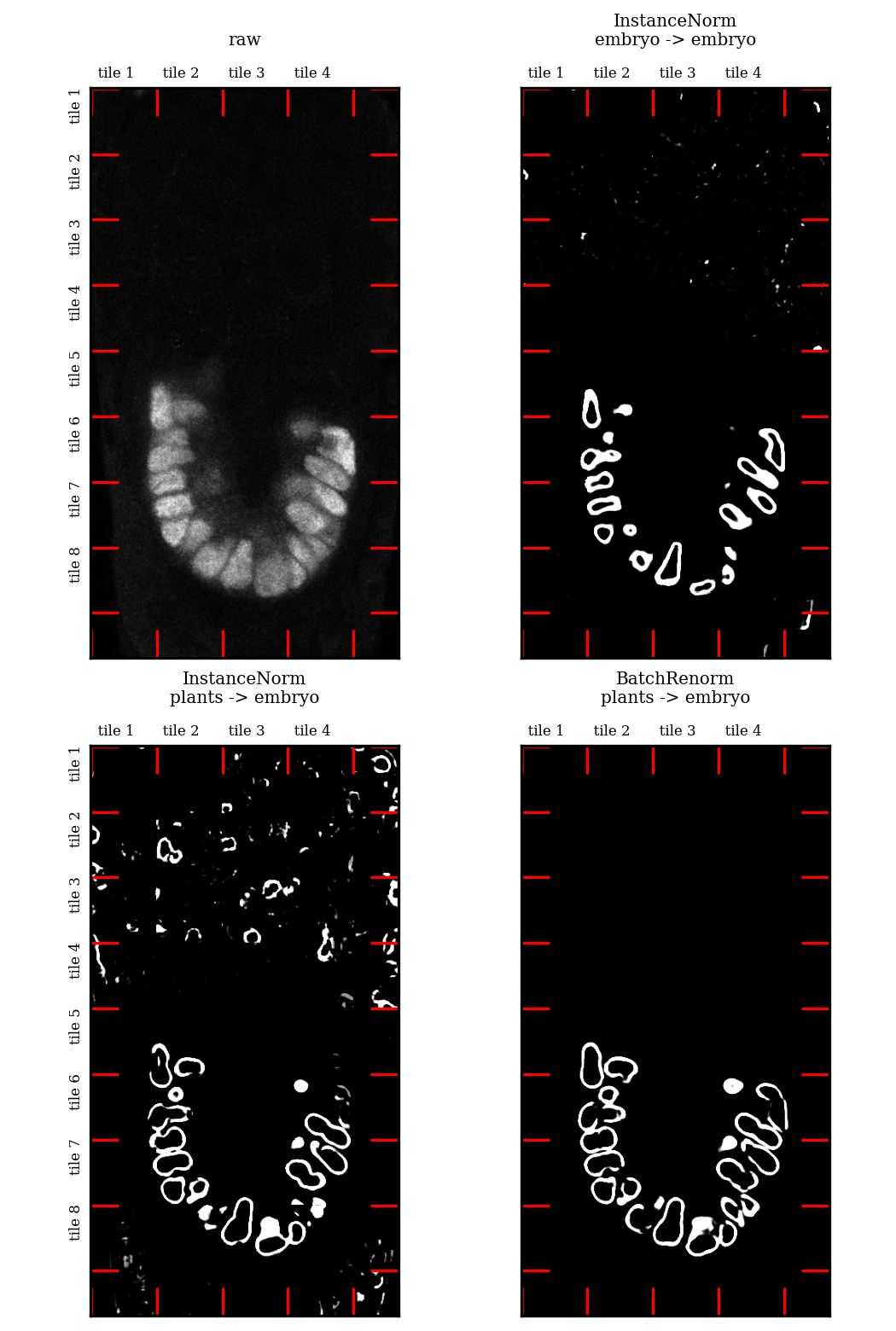}

   \caption{Tiling artifacts become more pronounced in a transfer setting. For quantitative evaluation see STab.~1.}
   \label{fig:transfer}
\end{figure}

\begin{figure}[t]
  \centering
    \includegraphics[width=\linewidth]{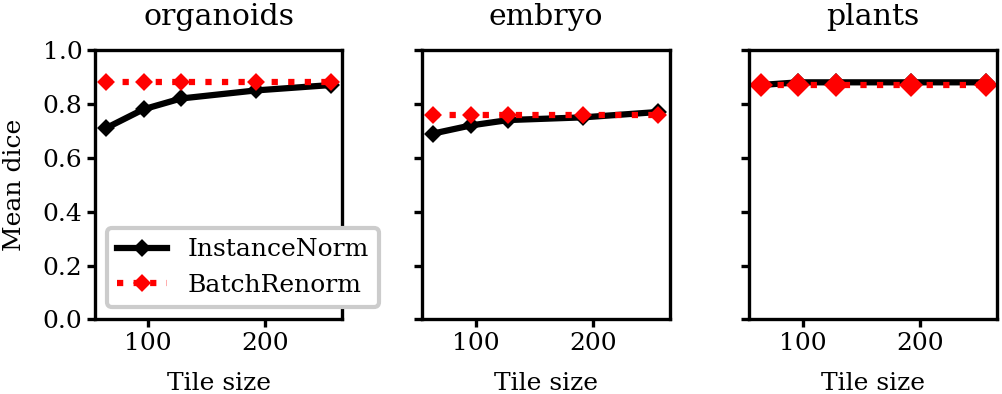}

   \caption{Performance vs tile size.}
   \label{fig:tilesize}
\end{figure}

\subsection{Receptive field}
 
We calculated TRF and ERF for the architectures with different normalization layers. As expected, for U-Net with the global feature normalization (\texttt{BatchNorm}, \texttt{BatchRenorm}, \texttt{Identity}) only a limited region around the central pixel belongs to TRF, ERF constitutes the effective receptive field only the small portion as shown in \cref{fig:receptive_field}. For the U-Net with tile-wise feature normalization (\texttt{InstanceNorm}) ERF is the same, which is expected since it is determined by the parameters of the convolutional architecture, but the TRF takes the whole tile as values of all pixels in the tile make a contribution to the prediction for the central pixel via normalization. Therefore, regardless of the size of the overlap between tiles in the sliding window prediction, the stitched prediction will not be exactly seamless. In all sliding window experiments with the U-Net we used halo calculated using TRF of the U-Net with global feature normalization (\cref{fig:receptive_field}).

For UNETR TRF also takes the whole input tile due to normalization in the convolutional decoder and the attention mechanism. The averaged gradients image \cref{fig:receptive_field} has a clear square pattern because in the encoder the image is downsampled by 16 and then upsampled back to the original resolution. Despite the attention layers, ERF is the same as for the U-Net. Since only 3 central patches have large effect on the central pixel prediction, we set $halo=32$ for all UNETR experiments (\cref{fig:receptive_field}).

\subsection{Tile-wise feature normalization with \texttt{InstanceNorm} causes tiling artifacts}

Using \texttt{InstanceNorm} allows to reach high quality predictions, however the tile-wise feature normalization is a source of artifacts both for U-Net (\cref{tab:unet}) and for UNETR (\cref{tab:unetr}) as shown in \cref{fig:artifact_correction}. We observe that tile-wise input normalization (\texttt{BatchNorm}, tile-wise input norm) causes less severe tile mismatch than tile-wise network feature normalization (\texttt{InstanceNorm}, global input norm). Although it is impossible to reach completely seamless stitching with UNETR, experiments with \texttt{InstanceNorm} show strong tile mismatch, while global feature normalization and sufficient halo based on ERF size allow to reduce mismatch and make it almost unnoticeable. 

Although \texttt{InstanceNorm} performs better than \texttt{BatchNorm} for boundaries, networks with tile-wise normalization struggle with classes which cover areas larger than the tile size, such as foreground and matrix in organoids dataset. Experiments with inference using different tile sizes \cref{fig:tilesize} show that the prediction depends strongly on the tile size, reaching the best quality when the tiles become large enough for the feature distribution in each tile to reproduce the distribution in the whole image. 

\cref{fig:transfer} shows how artifacts get worse in the transfer setting: the model originally trained on the embryo dataset produces less artifacts than the model transferred from the plants dataset. We observe the same effect in transfer in both directions (STab.~1). 

\renewcommand{\arraystretch}{1.45}
\begin{table}
\footnotesize
\centering

\caption{UNETR results table}
\label{tab:unetr}
\setlength{\tabcolsep}{2pt}
\begin{tabular}{c|c|c||l|l|l|l}

\toprule
\multicolumn{3}{l||}{Norm layer} & BatchNorm &  InstanceNorm & BatchRenorm & Identity \\

\multicolumn{3}{l||}{Input norm} & global & global & global & global  \\

\multicolumn{3}{l||}{Feature norm} & global & tile-wise & global & global  \\
\midrule
\midrule
\multirow{7}{*}{\rotatebox[origin=c]{90}{dice, eval mode}} & \multirow{3}{*}{\rotatebox[origin=c]{90}{organoids}}
& foreground & 0.93 & 0.89 & \textbf{0.95} & 0.91 \\
& & boundaries & 0.39 & 0.60 & \textbf{0.72} & 0.60 \\
& & matrix & 0.03 & 0.10 & \textbf{0.87} & 0.28 \\
\cline{2-7}
& \multirow{2}{*}{\rotatebox[origin=c]{90}{plants}}
 & foreground & \textbf{0.84} & \textbf{0.85} & \textbf{0.85} & 0.52 \\
& & boundaries & 0.64 & \textbf{0.69} & 0.66 & 0.21 \\
\cline{2-7}
& \multirow{2}{*}{\rotatebox[origin=c]{90}{embryo}}
 & foreground & \textbf{0.67} & 0.56 & \textbf{0.67} & 0.00 \\
& & boundaries & 0.49 & 0.49 & \textbf{0.53} & 0.00 \\
\midrule
\midrule
\multirow{7}{*}{\rotatebox[origin=c]{90}{dice, train mode}} & \multirow{3}{*}{\rotatebox[origin=c]{90}{organoids}}
& foreground & 0.89 & 0.89 & \textbf{0.95} & 0.91 \\
& & boundaries & 0.60 & 0.60 & \textbf{0.72} & 0.60 \\
& & matrix & 0.10 & 0.10 & \textbf{0.87} & 0.28 \\
\cline{2-7}
& \multirow{2}{*}{\rotatebox[origin=c]{90}{plants}}
 & foreground & \textbf{0.85} & \textbf{0.85} & \textbf{0.85} & 0.52 \\
& & boundaries & \textbf{0.69} & \textbf{0.69} & 0.66 & 0.21 \\
\cline{2-7}
& \multirow{2}{*}{\rotatebox[origin=c]{90}{embryo}}
 & foreground & 0.56 & 0.56 & \textbf{0.67} & 0.00 \\
& & boundaries & 0.49 & 0.49 & \textbf{0.53} & 0.00 \\

\midrule
\midrule
\multirow{3}{*}{\rotatebox[origin=c]{90}{t/e disparity}} & \multicolumn{2}{l||}{organoids} & 0.36 & \textbf{0} & \textbf{0} & \textbf{0} \\
\cline{2-7}
& \multicolumn{2}{l||}{plants} & 0.20 & \textbf{0} & \textbf{0} & \textbf{0} \\
\cline{2-7}
& \multicolumn{2}{l||}{embryo} &  0.33 & \textbf{0} & \textbf{0} & \textbf{0} \\

\midrule
\midrule
\multirow{3}{*}{\rotatebox[origin=c]{90}{tile mismatch}} & \multicolumn{2}{l||}{organoids} & \textbf{0.00} & 0.10 & \textbf{0.00} & \textbf{0.00} \\
\cline{2-7}
& \multicolumn{2}{l||}{plants} & \textbf{0.00} & 0.04 & \textbf{0.00} & \textbf{0.00} \\
\cline{2-7}
& \multicolumn{2}{l||}{embryo} & \textbf{0.00} & 0.06 & \textbf{0.00} & \textbf{0.00} \\

\midrule
\bottomrule
\end{tabular}
\end{table}

\subsection{Mismatch in batch statistics causes \texttt{train} and \texttt{eval} mode disparity for \texttt{BatchNorm}}
\begin{figure}[t]
  \centering

    \includegraphics[width=\linewidth]{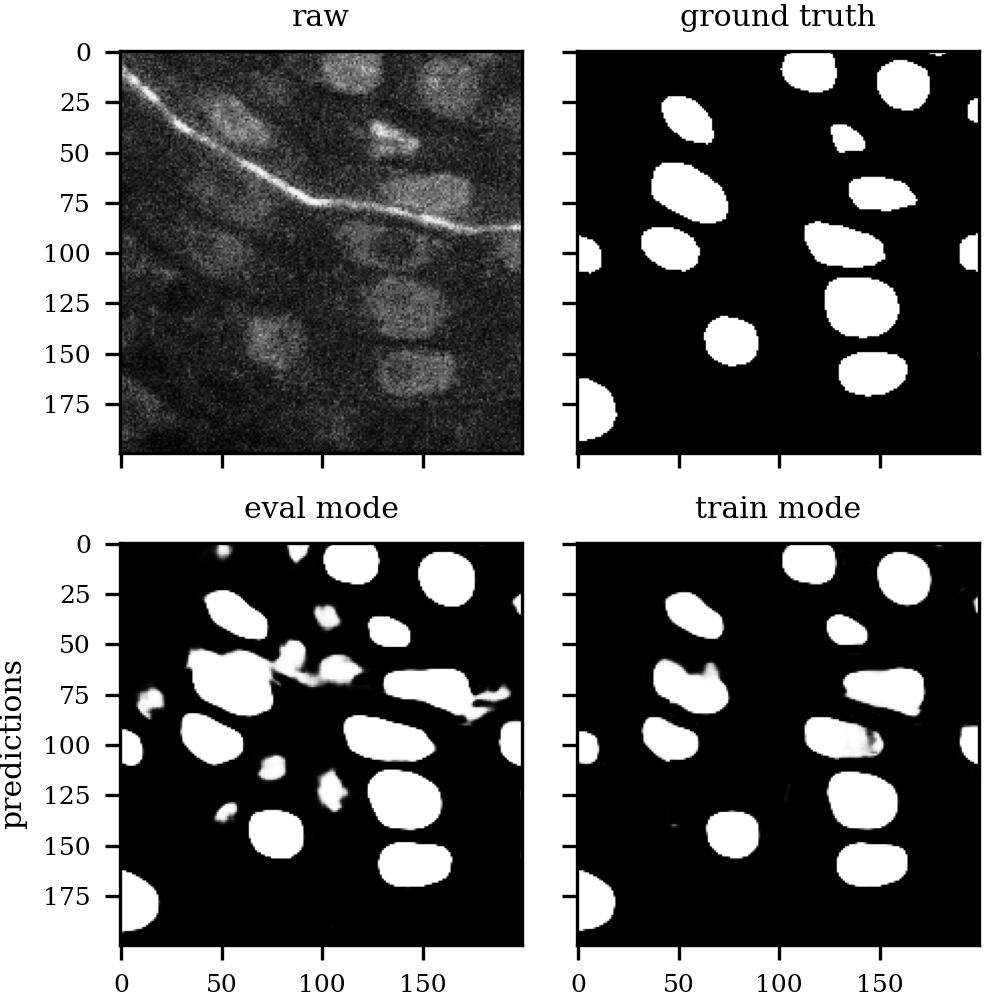}

   \caption{Example of train/eval disparity}
   \label{fig:train_eval_disparity}
\end{figure}

Global feature normalization with \texttt{BatchNorm} allows to eliminate the tile mismatch but predictions for organoids boundary and other datasets are worse than with \texttt{InstanceNorm}. The reason for this is the disparity between the performance of the network in train and evaluation mode. Estimation of normalization statistics with a running average in \texttt{BatchNorm} relies on the batch statistics being stable enough to represent the whole dataset. If batches are too small, the running average can be very different from the batch-wise statistics used during training, leading to deteriorating performance. Doing inference with the network in train mode supports this hypothesis: the performance recovers, at the cost of tiling artifacts. In our experiments train and evaluation mode predictions can be quite different, with train/eval disparity reaching $0.84$ for the organoids dataset. Qualitatively, the changes in the predictions occur in the areas of low prediction certainty. For example, in \cref{fig:train_eval_disparity} the eval mode differs from the train mode in the areas where the background has a bright line artifact or in the noisier areas.

This effect is not the same as the usual difference between train and evaluation performance caused by overfitting. Here, the network's output differs even for the exact same images thus violating the usual assumption that the prediction quality is approximately the same in train and eval mode. This issue would not be remedied by using augmentations or more training data as it does not address the high variance of statistics.

\subsection{BatchRenorm corrects for train/eval disparity and provides seamless stitching}
\texttt{InstanceNorm} avoids the statistics mismatch by always using the statistics of the input both during training and during inference at the cost of causing tiling artifacts. \texttt{BatchNorm} does not introduce the tiling artifacts but can lose performance due to train/eval disparity. 

A potential solution is to remove the normalization altogether. In our experiments not using normalization (\texttt{Identity} in \cref{tab:unet} and \cref{tab:unetr}) has led to worse performance for plants dataset and similar performance for organoids and embryo. It could be possible to improve the performance of networks with no normalization by finding more suitable training parameters and reducing the learning rate, however the necessity to adjust training parameters for each dataset makes the pipeline harder to use.

Another solution is to use the same global normalization statistics both in training and in inference. Directly using running average statistics during training causes network's weights to explode \cite{BatchRenorm}, which is why we propose to use Batch Renormalization method. We found that networks with \texttt{BatchRenorm} produce same or better predictions as \texttt{InstanceNorm} (\cref{tab:unet}, \cref{tab:unetr}) including the transfer setup (STab. 1) while allowing for seamless stitching due to global feature normalization (last column of \cref{fig:artifact_correction},  \cref{fig:transfer}).

\section{Discussion}
\label{sec:discussion}
The severity of tiling artifacts depends on the data and on the tile size. For example, the tile mismatch can be negligible in case the images are very homogeneous in content and the tile size is big enough that each tile mostly reproduces the global statistics of the image, such as for plants dataset in our experiments. However, even in that case the inference performance depends on the tile size and can change unpredictably in the transfer setting, putting higher requirements on the hardware for inference and making the pipeline harder to use, especially for non-expert users.  Tracing low validation accuracy to the instability in batch statistics is not straightforward either, as it is normally expected that the performance on the validation dataset should be worse than on the training data. If the training loss is low but the accuracy on validation data does not improve, it is often attributed to the insufficient amount of ground truth, however, we show that it can also be caused by train/eval disparity preventing efficient utilization of the ground truth. 

Both of the described effects are highly dependent on the particular data, input normalization, sampling strategy, postprocessing and other parameters of the image analysis pipeline. For any such pipeline, we introduce two simple metrics: $tile\ mismatch$ and $train/eval\ disparity$ to help determine if it has issues with normalization. Both metrics are easy to use for any sliding window inference pipeline, including other dense prediction tasks if the Dice score is replaced with another, task-appropriate metric.

We found that \texttt{BatchRenorm}, originally introduced for natural image classification, performs well for segmenting large biological images. The challenge with images larger than GPU memory is ensuring consistent normalization parameters across the entire image for seamless stitching, despite the inability to process the full image during training due to time constraints. \texttt{BatchRenorm} solves this by splitting training into two stages. First, the network is trained with \texttt{BatchNorm}. Then the normalization layer switches to using the accumulated running average statistics. In our experience it is very important that the training converges before switching and that the statistics change from tile-wise to global gradually to achieve the best final accuracy. While the \texttt{BatchRenorm} adds more hyperparameters, in our experiments the right moment to switch could easily be found from looking at the loss curves.

\section{Conclusion}
\label{sec:conclusion}
In this paper, we demonstrate that tiling artifacts in sliding window inference pipelines can be caused by the tile-wise feature normalization in the network. We also find that  switching from tile-wise normalization in training to global normalization in evaluation can lead to even worse performance, including hallucinations, due to small batch size. As these effects strongly depend on the data and tile size, we introduce $tile\ mismatch$ and $train/eval\ disparity$ metrics to easily check if a specific pipeline and dataset have these issues. We propose that Batch Renormalization technique will yield artifact-free predictions even with small batch size and demonstrate its good performance through extensive experiments on various microscopy datasets and network architectures.

\textbf{Acknowledgment.} We would like to thank Ricardo Sánchez Loayza for helpful discussions. E.B. was supported by the Joachim Herz Foundation through an Add-on Fellowship for Interdisciplinary Life Science. E.D’I. was supported by a fellowship from the EMBL Interdisciplinary (EI4POD) program under Marie Sk1odowska-Curie Actions COFUND (847543). J.M. acknowledges funding from the EMBL and the European Research Council (ERC 3DCellPhase- 760067). This work was supported by the European Commission through the Horizon Europe program (IMAGINE project, grant agreement 101094250-IMAGINE). The work of A.A. was funded by the German Research Foundation through  PA 4341/2-1. C.P. received support under Germany’s Excellence Strategy - EXC 2067/1-390729940. We would like to thank the EMBL IT Services Department for providing computational infrastructure and support.

{
    \small
    \bibliographystyle{ieeenat_fullname}
    \bibliography{main}
}

\newpage
\appendix
\title{\center {\Large Supplementary materials}}

\section{Implementation details}
All experiments were implemented using MONAI\footnote{\hyperlink{https://monai.io}{https://monai.io}} framework. The models were trained on a slurm cluster with NVIDIA GeForce RTX 3090 GPUs with the batch size of 1 and gradient accumulation for 8 steps, using the Adam optimizer with initial learning rate of 0.001 for 25,000 iterations. The only augmentation was flip with probability 0.5. During training tiles of size $192 \times 192 \times 192$ were sampled from the volumes randomly. Dice loss averaged over channels was used for training. Dice is not defined when the label has no non-zero pixels, therefore for tiles with empty channels the empty channels were not used in averaging. 

U-Net with two downsampling steps and 32, 64 and 128 feature maps in the first, second and bottleneck levels was used. For all normalization layers that use running average to collect statistics \texttt{momentum} was set to 0.01. 

UNETR encoder had $16\times16\times16$ patch size, 12 attention heads and hidden embedding size 768. Convolutional patch projection and trainable positional embedding were used, following the recommendations in the UNETR paper.

For the Batch Renormalization layer (adapted from https://github.com/ludvb/batchrenorm) we set $r_{max}=3$, $d_{max}=5$ following the original paper. The models were trained for $1000$ steps for embryo and organoids datasets and $5000$ steps for plants dataset with $r=1$ and $d=0$, then the parameters were linearly increased to $r_{max}$ and $d_{max}$ over $1000$ steps.

\setcounter{table}{0}
\renewcommand{\arraystretch}{1.6}
\begin{table*}
\small
\centering
\caption{U-Net transfer between plants and embryo datasets. Values in gray correspond to the non-transfer experiments.}
\label{tab:transfer}
\setlength{\tabcolsep}{2pt}
\begin{tabular}{c|c|c||cc|cc|cc|cc|cc|cc}

\toprule
\multicolumn{3}{l||}{Normalization layer} & \multicolumn{2}{p{1.6cm}|}{BatchNorm} & \multicolumn{2}{p{2cm}}{BatchNorm} & \multicolumn{2}{p{2cm}|}{InstanceNorm} & \multicolumn{2}{p{2cm}|}{InstanceNorm} & \multicolumn{2}{p{2cm}|}{BatchRenorm} & \multicolumn{2}{p{2cm}|}{Identity} \\

\multicolumn{3}{l||}{Input norm} & \multicolumn{2}{l|}{global} & \multicolumn{2}{l|}{tile-wise} & \multicolumn{2}{l|}{global} & \multicolumn{2}{l|}{tile-wise} & \multicolumn{2}{l|}{global} & \multicolumn{2}{l|}{global}  \\

\multicolumn{3}{l||}{Feature norm} & \multicolumn{2}{l|}{global} & \multicolumn{2}{l|}{global} & \multicolumn{2}{l|}{tile-wise} & \multicolumn{2}{l|}{tile-wise} & \multicolumn{2}{l|}{global} & \multicolumn{2}{l|}{global}  \\
\midrule
\multicolumn{3}{l||}{\backslashbox[3.3cm]{Target}{Source}} & plant & embryo & plant & embryo & plant & embryo  & plant & embryo & plant & embryo & plant & embryo\\
\midrule
\midrule

\multirow{4}{*}{\rotatebox[origin=c]{90}{dice, eval mode}} & \multirow{2}{*}{\rotatebox[origin=c]{90}{plants}}
 & foreground & \textcolor{gray}{0.85} & 0.36 & \textcolor{gray}{0.88} & 0.42 & \textcolor{gray}{0.89} & 0.50 & \textcolor{gray}{0.89} & \textbf{0.53} & \textcolor{gray}{0.89} & 0.50 & \textcolor{gray}{0.52} & 0.21 \\
     
& & boundaries & \textcolor{gray}{0.69} & 0.11 & \textcolor{gray}{0.73} & 0.14 & \textcolor{gray}{0.75} & 0.27 & \textcolor{gray}{0.74} & 0.22 & \textcolor{gray}{0.74} & \textbf{0.29} & \textcolor{gray}{0.26} & 0.25 \\
\cline{2-15}
& \multirow{2}{*}{\rotatebox[origin=c]{90}{embryo}}
 & foreground &  0.51 & \textcolor{gray}{0.66} & 0.51 & \textcolor{gray}{0.68} & 0.54 & \textcolor{gray}{0.67} & 0.54 & \textcolor{gray}{0.72} & \textbf{0.59} & \textcolor{gray}{0.74} & 0.45 & \textcolor{gray}{0.59} \\
& & boundaries & 
0.18 & \textcolor{gray}{0.55} & 0.14 & \textcolor{gray}{0.55} & 0.28 & \textcolor{gray}{0.55} & 0.28 & \textcolor{gray}{0.53} & \textbf{0.29} & \textcolor{gray}{0.58} & 0.12 & \textcolor{gray}{0.44} \\
\midrule
\midrule
\multirow{2}{*}{\rotatebox[origin=c]{90}{tile mism.}} & \multicolumn{2}{l||}{plants} & \textcolor{gray}{\textbf{no}} & \textbf{no} & \textcolor{gray}{0.01} & 0.00 & \textcolor{gray}{0.05} & 0.10 & \textcolor{gray}{0.05} & 0.07 & \textcolor{gray}{\textbf{no}} & \textbf{no} & \textcolor{gray}{\textbf{no}} & \textbf{no} \\
\cline{2-15}
& \multicolumn{2}{l||}{embryo} & \textbf{no} & \textcolor{gray}{\textbf{no}} & 0.17 & \textcolor{gray}{0.09} & 0.18 & \textcolor{gray}{0.16} & 0.25 & \textcolor{gray}{0.20} & \textbf{no} & \textcolor{gray}{\textbf{no}} & \textbf{no} & \textcolor{gray}{\textbf{no}} \\

\midrule
\bottomrule
\end{tabular}
\end{table*}

\end{document}